\pdfoutput=1
\documentclass[11pt]{article}

\usepackage{ACL2023}

\usepackage{times}
\usepackage{latexsym}
\usepackage[most]{tcolorbox}
\usepackage{amsmath}
\usepackage{pifont}
\usepackage{multirow}
\usepackage{soul}
\usepackage{booktabs}
\usepackage{amssymb}
\usepackage{adjustbox}
\usepackage{subcaption}
\usepackage{xcolor}
\usepackage{xspace}
\usepackage{algorithm}
\usepackage{algpseudocode}
\newcommand{\stitle}[1]{\vspace{1ex} \noindent{\bf #1.}}

\newcommand{\gpttwo}{\textsc{GPT-2}\xspace}
\newcommand{\gptthree}{\textsc{GPT-3}\xspace}
\newcommand{\gptthreepointfive}{\textsc{GPT-3.5}\xspace}
\newcommand{\gptfour}{\textsc{GPT-4}\xspace}
\newcommand{\gptfouro}{\textsc{GPT-4o}\xspace}
\newcommand{\gptfouromini}{\textsc{GPT-4o-mini}\xspace}
\newcommand{\llamaone}{\textsc{Llama3-8B}\xspace}
\newcommand{\llamatwo}{\textsc{Llama3-70B}\xspace}
\newcommand{\llamathree}{\textsc{Llama3.1-8B}\xspace}
\newcommand{\llamafour}{\textsc{Llama3.1-70B}\xspace}
\newcommand{\gemmaone}{\textsc{Gemma2-9B}\xspace}
\newcommand{\gemmatwo}{\textsc{Gemma2-27B}\xspace}
\newcommand{\qwenone}{\textsc{Qwen1.5-72B}\xspace}
\newcommand{\qwentwo}{\textsc{Qwen2-72B}\xspace}

\newcommand{\gpt}{\textsc{GPT}\xspace}
\newcommand{\llama}{\textsc{Llama}\xspace}

\newcommand{\llamathreeseries}{\textsc{Llama3}\xspace}
\newcommand{\gemma}{\textsc{Gemma}\xspace}
\newcommand{\gemmatwoseries}{\textsc{Gemma2}\xspace}
\newcommand{\qwen}{\textsc{Qwen}\xspace}

\usepackage[T1]{fontenc}
\usepackage[utf8]{inputenc}

\usepackage{microtype}

\usepackage{inconsolata}

\tcbset{
    userstyle/.style={
        enhanced,
        colback=white,
        colframe=black,
        colbacktitle=gray!20,
        coltitle=black,
        rounded corners,
        sharp corners=north,
        boxrule=0.5pt,
        drop shadow=black!50!white,
        attach boxed title to top left={
            xshift=-2mm,
            yshift=-2mm
        },
        boxed title style={
            rounded corners,
            size=small,
            colback=gray!20
        }
    },
    replystyleg/.style={
        enhanced,
        colback=green!15,
        colframe=black,
        colbacktitle=green!30,
        coltitle=black,
        boxrule=0.5pt,
        drop shadow=black!50!white,
        rounded corners,
        sharp corners=north,
        attach boxed title to top right={
            xshift=-2mm,
            yshift=-2mm
        },
        boxed title style={
            rounded corners,
            size=small,
            colback=green!40
        }
    },
    replystyler/.style={
        enhanced,
        colback=red!15,
        colframe=black,
        colbacktitle=red!40,
        coltitle=black,
        boxrule=0.5pt,
        drop shadow=black!50!white,
        rounded corners,
        sharp corners=north,
        attach boxed title to top right={
            xshift=-2mm,
            yshift=-2mm
        },
        boxed title style={
            rounded corners,
            size=small,
            colback=red!40
        }
    }
}

\newtcolorbox{userquery}[1][]{
    userstyle,
    title=Prompt,
    #1
}

\definecolor{darkred}{RGB}{200,0,0}
\definecolor{lightgreen}{RGB}{160,230,160}
\definecolor{darkgreen}{RGB}{0, 100, 0}
\definecolor{lightred}{RGB}{252,231,234}
\definecolor{lightyellow}{RGB}{250,253,191}
\definecolor{lightblue}{RGB}{230,240,254}
\definecolor{lightorange}{RGB}{255,223,191}
\definecolor{white}{RGB}{255,255,255}

\newcommand\hlc[2]{\sethlcolor{#1} \hl{#2}}

\newcommand{\greentext}[1]{{\hlc{lightgreen}{#1}}}

\title{Vulnerability of LLMs to Vertically Aligned Text Manipulations}

\author{
Zhecheng Li$^\dagger$ \ \ \ \ Yiwei Wang$^{\ddagger \delta}$ \ \ \ \ Bryan Hooi$^\|$ \ \ \ \ Yujun Cai$^\mathsection$ \ \ \ \ Zhen Xiong$^\mathparagraph$ \\ 
\ \ \ \ \textbf{Nanyun Peng}$^\ddagger$ \ \ \ \ \textbf{Kai-Wei Chang}$^\ddagger$ \\
$^\dagger$ University of California, San Diego \quad $^\ddagger$ University of California, Los Angeles \\
$^\mathsection$ The University of Queensland \quad $^\|$ National University of Singapore \\
$^\mathparagraph$ University of Southern California \quad $^\delta$ University of California, Merced \\
\texttt{zhl186@ucsd.edu}
}

\begin{document}
\maketitle
\begin{abstract}

Vertical text input is commonly encountered in various real-world applications, such as mathematical computations and word-based Sudoku puzzles. While current large language models (LLMs) have excelled in natural language tasks, they remain vulnerable to variations in text formatting.
Recent research demonstrates that modifying input formats, such as vertically aligning words for encoder-based models, can substantially lower accuracy in text classification tasks. While easily understood by humans, these inputs can significantly mislead models, posing a potential risk of bypassing detection in real-world scenarios involving harmful or sensitive information.
With the expanding application of LLMs, a crucial question arises: \textit{Do decoder-based LLMs exhibit similar vulnerabilities to vertically formatted text input?} In this paper, we investigate the impact of vertical text input on the performance of various LLMs across multiple text classification datasets and analyze the underlying causes. Our findings are as follows: (i) Vertical text input significantly degrades the accuracy of LLMs in text classification tasks. (ii) \textit{Chain of Thought (CoT)} reasoning does not help LLMs recognize vertical input or mitigate its vulnerability, but \textit{few-shot learning} with careful analysis does. (iii) We explore the underlying cause of the vulnerability by analyzing the inherent issues in tokenization and attention matrices.
\end{abstract}

\section{Introduction}
Text classification is one of the most common tasks in Natural Language Processing (NLP), encompassing a wide range of applications, including sentiment analysis, harmful content detection, and spam filtering~\citep{textclassification1, textclassification2, textclassification3, textclassification4, textclassification5}. Since the introduction of the Transformer architecture~\citep{transformer}, models based on this architecture, such as BERT models~\citep{bert, distilbert} and \gpt models~\citep{gpt2, gpt3, gpt4}, have achieved impressive performance in text classification~\citep{transformertc1, transformertc2, transformertc3}. These models typically operate by either fine-tuning encoder-based Transformers~\citep{bertfinetune1, bertfinetune2, bertfinetune3, bertfinetune4} or leveraging decoder-based models to generate text outputs directly~\citep{llmtc1, llmtc2, llmtc3}. Their success across various text classification tasks has made them indispensable tools in NLP.

\begin{figure}[!tb]
    \centering
    \includegraphics[width=1.00\linewidth]{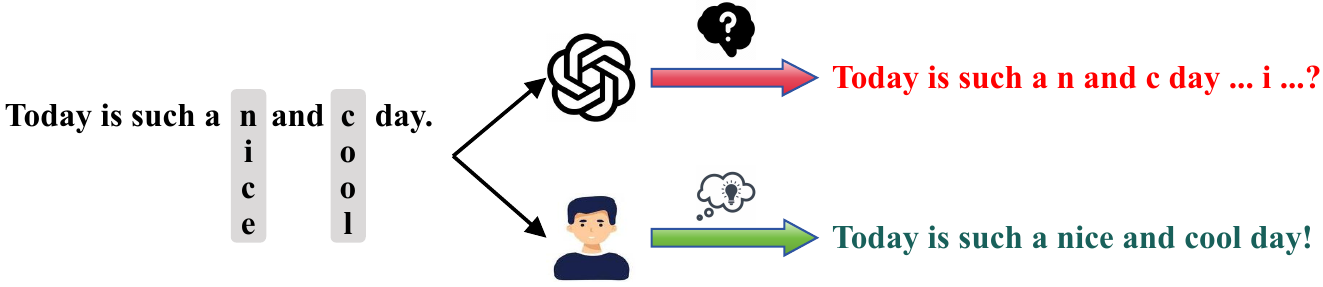}
    \caption{Humans can easily comprehend vertically formatted input by mentally processing the transformation, models often struggle to interpret it.
    \label{fig:example}}
    \vspace{-2mm}
\end{figure}

However, despite these advances, language models remain sensitive to certain types of input variation. Seemingly minor formatting changes, such as line breaks, punctuation marks, or word order, can significantly affect model outputs~\citep{modelsensitive, professor}. A particularly intriguing example of this is vertical text formatting, where words are arranged vertically rather than horizontally. Although this format poses no challenge for human readers, it can confuse language models (shown in Figure \ref{fig:example}). Meanwhile, vertical text finds practical applications in areas such as representing tree structures in computing, bypassing social media detection, and even educational games, such as Sudoku word and acrostic poetry. Therefore, understanding how language models interpret vertical text is essential for enhancing their utility.

As the capabilities of large language models (LLMs) such as the \gpt series~\citep{gpt2, gpt3, gpt4} and \llama series~\citep{llama3, llama2} continue to expand, evaluating their ability to understand vertical input becomes increasingly important. Although some of the research on input sensitivity has focused on encoder-based models such as BERT~\citep{bert, distilbert}, the behavior of LLMs when faced with unconventional input formats, such as vertical text, remains underexplored. Given that LLMs are increasingly being applied in critical areas such as content moderation, spam filtering, and misinformation detection, understanding their ability to handle atypical input is crucial to ensure robustness in real-world applications~\citep{llmspam, llmtoxic1, llmtoxic2}.

To address this gap, we systematically evaluate the impact of vertical text formatting on LLM performance across various text classification tasks. We hypothesize that, despite their advanced capabilities, LLMs still exhibit similar vulnerabilities to vertically formatted input as encoder-based models.

To explore this, we conduct experiments on various open-source and closed-source LLMs across various text classification tasks. Our results show a significant decrease in model accuracy when exposed to vertical text input, indicating that even state-of-the-art LLMs struggle with this input format. We further analyze tokenization and attention matrices to investigate the underlying mechanisms causing this degradation. Additionally, we explore mitigation strategies, finding that few-shot learning can effectively improve the models’ ability to handle vertical text, while CoT reasoning does not provide similar benefits.

\begin{figure*}[!tb]
	\centering
	\includegraphics[width=1.00\linewidth]{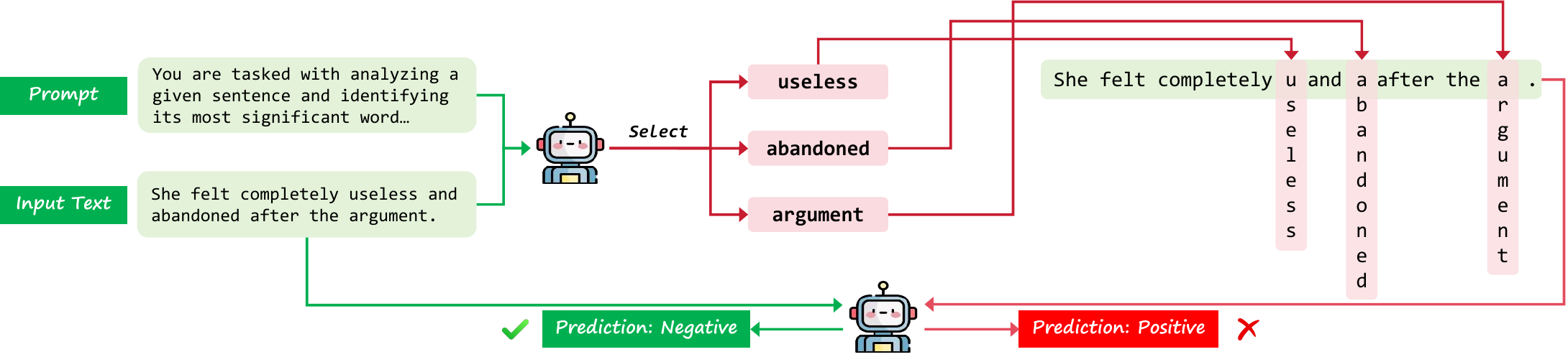}
	\caption{An example of the entire algorithm process demonstrates that vertically inputted words lead the LLM to incorrectly predict the label for text classification.
	\label{fig:algorithm}}
    \vspace{-2mm}
\end{figure*}

In summary, our contributions are as follows:
\begin{itemize}
    \item We present the first comprehensive study on the vulnerability of LLMs to vertically formatted text, demonstrating that this formatting significantly impairs text classification performance.
    \item Our analysis of tokenization and attention patterns provides insight into the underlying reasons for LLM performance degradation when handling vertical input.
    \item We propose a few-shot learning with a manually crafted analysis as an effective strategy to mitigate the impact of vertical text formatting, while CoT reasoning proves to be less effective in this context.
\end{itemize}

\section{Related Works}
The introduction of the transformer architecture marked the beginning of the LLM era in natural language processing~\citep{transformer}. This evolution started with early models such as \gpttwo~\citep{gpt2} and \gptthree~\citep{gpt3} and advanced to more semantically capable and safer models such as ChatGPT, \gptfour, and \gptfouro~\citep{gpt4}. These models have been further refined through post-training techniques, including RLHF~\citep{rlhf} and DPO~\citep{dpo}. Throughout their development, concerns and research on the vulnerability and susceptibility of LLMs to various attacks have remained a key focus.

For example, in recent years, researchers have developed various methods to `jailbreak' LLMs, causing them to generate harmful content~\citep{jailbreaksurvey, jailbreak2, jailbreak3, jailbreak4}. Such attacks clearly highlight the inherent vulnerabilities of LLMs. However, these vulnerabilities extend beyond this issue. Studies have shown that LLMs can produce significantly different outputs with similar input texts depending on content order (e.g., in multiple choice questions)~\citep{professor}. Moreover, LLMs often show pronounced sensitivity to basic punctuation and line breaks, further illustrating their fragility~\citep{modelsensitive}.

Recent research reveals that vertically formatted text can severely impair the comprehension abilities of Transformer encoder-based models, such as BERT, causing significant performance degradation~\citep{vertattack, bert}. This vulnerability, where straightforward text can substantially deceive language models, raises concerns about the potential impact of such formatting on contemporary LLMs. If LLMs are similarly susceptible to vertically formatted text, this could pose a threat to real-world applications.

\section{Methodology}
To assess the vulnerability of LLMs to vertically formatted text, we will select a set of semantically significant words from a given text and input them vertically into the LLMs. The remaining words will be presented in the standard horizontal format from left to right. This approach mirrors real-world usage, where most text is input normally to ensure coherence and readability, while the vertically formatted keywords test the model's susceptibility. The method comprises two main components: \textit{Word Selection} and \textit{Word Transformation}, as shown in Figure \ref{fig:algorithm}.

\subsection{Word Selection}
In previous work, researchers used a greedy approach to assess the impact of each word on the prediction probability of the model, thus determining the importance of individual words in a sentence~\citep{vertattack}. However, this method is time-consuming and resource-intensive for LLMs. In this paper, we address this issue by leveraging prompt-based LLMs as evaluators to identify and extract key words from the text. 
% We show our prompts in Appendix \ref{sec:wordselectionprompts}.

\subsection{Word Transformation}
The goal of word transformation is to vertically arrange specific words within a sentence. The workflow is as follows:

\stitle{Decompose the Sentence} The input sentence is divided into a list of words, and the height of the vertical arrangement is determined by the longest designated word.

\stitle{Initialize the Grid} We create a two-dimensional grid based on the length of the longest word, initially filled with empty spaces to hold the characters of the vertical words.

\stitle{Align Vertical Characters} Each vertical word's characters are placed in corresponding grid rows, with shorter words padded to maintain alignment.

\stitle{Handle Non-Vertical Words} Characters of non-vertical words are placed in their respective rows, with remaining spaces filled to ensure formatting consistency.

\stitle{Generate the Final String} Each grid row is concatenated into a string, separated by newlines and adjusted for alignment, resulting in a layout that displays vertical and horizontal words as intended.

\begin{table*}[!tb]
	\centering
	\begin{adjustbox}{width=1.00\linewidth}
    \renewcommand{\arraystretch}{1.30}
        \setlength{\tabcolsep}{3pt}
        \resizebox{\linewidth}{!}{
    		\begin{tabular}{@{}lcccccccccc@{}}
    		\toprule
            \multirow{2}{*}{\textbf{Model}} & \multicolumn{2}{c}{\textbf{SST-2}} & \multicolumn{2}{c}{\textbf{CoLA $^\spadesuit$}} & \multicolumn{2}{c}{\textbf{QNLI}} & \multicolumn{2}{c}{\textbf{Rotten Tomatoes}} & \multicolumn{2}{c}{\textbf{Jigsaw Toxicity}} \\ \cmidrule{2-11} 
             & \textbf{Original} & \textbf{Vertical} & \textbf{Original} & \textbf{Vertical} & \textbf{Original} & \textbf{Vertical} & \textbf{Original} & \textbf{Vertical} & \textbf{Original} & \textbf{Vertical} \\ \midrule \midrule
            \multicolumn{11}{c}{\textbf{Closed-Source Models}} \\ \midrule 
            \gptthreepointfive & $93.00$ & $65.00$ $(\downarrow\textcolor{red}{28.00})$ & $80.00$ & $47.00$ $(\downarrow\textcolor{red}{33.00})$ & $85.00$ & $69.00$ $(\downarrow\textcolor{red}{16.00})$ & $92.00$ & $57.00$ $(\downarrow\textcolor{red}{35.00})$ & $85.00$ & $62.00$ $(\downarrow\textcolor{red}{23.00})$ \\
            \gptfour & $96.00$ & $67.00$ $(\downarrow\textcolor{red}{29.00})$ & $90.00$ & $49.00$ $(\downarrow\textcolor{red}{41.00})$ & $89.00$ & $71.00$ $(\downarrow\textcolor{red}{18.00})$ & $93.00$ & $64.00$ $(\downarrow\textcolor{red}{29.00})$ & $89.00$ & $58.00$ $(\downarrow\textcolor{red}{31.00})$ \\
            \gptfouromini & $95.00$ & $66.00$ $(\downarrow\textcolor{red}{29.00})$ & $89.00$ & $50.00$ $(\downarrow\textcolor{red}{39.00})$ & $90.00$ & $71.00$ $(\downarrow\textcolor{red}{19.00})$ & $91.00$ & $61.00$ $(\downarrow\textcolor{red}{30.00})$ & $85.00$ & $57.00$ $(\downarrow\textcolor{red}{28.00})$ \\
            \gptfouro & $95.00$ & $68.00$ $(\downarrow\textcolor{red}{27.00})$ & $87.00$ & $47.00$ $(\downarrow\textcolor{red}{40.00})$ & $90.00$ & $70.00$ $(\downarrow\textcolor{red}{20.00})$ & $90.00$ & $65.00$ $(\downarrow\textcolor{red}{25.00})$ & $91.00$ & $60.00$ $(\downarrow\textcolor{red}{31.00})$ \\ \midrule
            \multicolumn{11}{c}{\textbf{Open-Source Models}} \\ \midrule
            \llamaone & $89.00$ & $61.00$ $(\downarrow\textcolor{red}{28.00})$ & $75.00$ & $50.00$ $(\downarrow\textcolor{red}{25.00})$ & $83.00$ & $62.00$ $(\downarrow\textcolor{red}{21.00})$ & $86.00$ & $42.00$ $(\downarrow\textcolor{red}{44.00})$ & $88.00$ & $58.00$ $(\downarrow\textcolor{red}{30.00})$ \\
            \llamatwo & $96.00$ & $67.00$ $(\downarrow\textcolor{red}{29.00})$ & $85.00$ & $48.00$ $(\downarrow\textcolor{red}{37.00})$ & $86.00$ & $63.00$ $(\downarrow\textcolor{red}{23.00})$ & $91.00$ & $46.00$ $(\downarrow\textcolor{red}{45.00})$ & $88.00$ & $58.00$ $(\downarrow\textcolor{red}{30.00})$ \\
            \llamathree & $93.00$ & $51.00$ $(\downarrow\textcolor{red}{42.00})$ & $80.00$ & $49.00$ $(\downarrow\textcolor{red}{31.00})$ & $83.00$ & $59.00$ $(\downarrow\textcolor{red}{24.00})$ & $89.00$ & $54.00$ $(\downarrow\textcolor{red}{45.00})$ & $80.00$ & $64.00$ $(\downarrow\textcolor{red}{16.00})$ \\
            \llamafour & $96.00$ & $66.00$ $(\downarrow\textcolor{red}{30.00})$ & $84.00$ & $50.00$ $(\downarrow\textcolor{red}{34.00})$ & $84.00$ & $66.00$ $(\downarrow\textcolor{red}{18.00})$ & $92.00$ & $63.00$ $(\downarrow\textcolor{red}{29.00})$ & $87.00$ & $62.00$ $(\downarrow\textcolor{red}{25.00})$ \\ \midrule
            \gemmaone & $88.00$ & $60.00$ $(\downarrow\textcolor{red}{28.00})$ & $83.00$ & $51.00$ $(\downarrow\textcolor{red}{32.00})$ & $78.00$ & $60.00$ $(\downarrow\textcolor{red}{18.00})$ & $86.00$ & $58.00$ $(\downarrow\textcolor{red}{28.00})$ & $85.00$ & $53.00$ $(\downarrow\textcolor{red}{32.00})$ \\
            \gemmatwo & $94.00$ & $58.00$ $(\downarrow\textcolor{red}{36.00})$ & $87.00$ & $50.00$ $(\downarrow\textcolor{red}{37.00})$ & $83.00$ & $64.00$ $(\downarrow\textcolor{red}{19.00})$ & $89.00$ & $54.00$ $(\downarrow\textcolor{red}{35.00})$ & $88.00$ & $55.00$ $(\downarrow\textcolor{red}{33.00})$ \\ \midrule
            \qwenone & $95.00$ & $63.00$ $(\downarrow\textcolor{red}{32.00})$ & $81.00$ & $52.00$ $(\downarrow\textcolor{red}{29.00})$ & $85.00$ & $66.00$ $(\downarrow\textcolor{red}{19.00})$ & $93.00$ & $57.00$ $(\downarrow\textcolor{red}{36.00})$ & $88.00$ & $63.00$ $(\downarrow\textcolor{red}{25.00})$ \\
            \qwentwo & $96.00$ & $60.00$ $(\downarrow\textcolor{red}{36.00})$ & $84.00$ & $50.00$ $(\downarrow\textcolor{red}{34.00})$ & $88.00$ & $62.00$ $(\downarrow\textcolor{red}{26.00})$ & $93.00$ & $59.00$ $(\downarrow\textcolor{red}{34.00})$ & $91.00$ & $59.00$ $(\downarrow\textcolor{red}{32.00})$ \\
            \bottomrule
    		\end{tabular}
            \vspace{-2mm}
        }
	\end{adjustbox}
	\caption{Accuracy scores of different LLMs on five datasets using zero-shot prediction, with changes before and after shown in parentheses (indicating the decrease). Dataset marked with $^\spadesuit$ includes two words inputted in a vertical format, while the others contain four words vertically inputted.
    \label{tab:mainexp}}
\end{table*}

\section{Experiments}
\subsection{Datasets}
We propose that vertically formatted text may disrupt LLMs' comprehension of content, potentially posing a threat by circumventing their detection mechanisms. In this paper, we therefore select five text classification datasets, covering a range of applications including sentiment analysis, harmful statement detection, and more:

\begin{itemize}
    \item SST-2~\citep{sst2}: A benchmark dataset for sentiment classification, featuring labeled movie reviews.
    \item CoLA~\citep{cola}: A dataset for evaluating linguistic acceptability, containing English sentences labeled as grammatically acceptable or unacceptable.
    \item QNLI~\citep{qnli}: Derived from the Stanford Question Answering Dataset (SQuAD), which evaluates whether a sentence answers a specific question.
    \item Rotten Tomatoes~\citep{rt}: A dataset of movie reviews used for sentiment analysis, with each review classified as positive or negative.
    \item Jigsaw Toxicity\footnote{\url{https://www.kaggle.com/c/jigsaw-toxic-comment-classification-challenge}}~\citep{jt}: A dataset of online comments labeled for various types of toxic behavior.
\end{itemize}

% \footnote{\url{https://www.kaggle.com/c/jigsaw-toxic-comment-classification-challenge}}

\subsection{Large Language Models}
To more effectively assess the vulnerability of latest LLMs to vertically formatted input text, we select various well-known models for experiments. 
% The introduction of different LLMs is shown in Appendix \ref{sec:llmintro}.

\stitle{Closed-Source Models} We utilize four different models developed by OpenAI, including the latest powerful models \gptfouro and \gptfouromini.

\stitle{Open-Source Models} We conduct our experiments on four models from the \llama series, two models from the \gemma series, and two models from the \qwen series, with parameter sizes ranging from $8$ billion to $72$ billion~\citep{llama3, gemma2, qwen2}.

\subsection{Metric}
We use the most straightforward metric: \textbf{Accuracy}. We evaluated the model by comparing its predicted labels with the actual labels. The accuracy of a model is calculated as follows:

\begin{equation}
    \text{Accuracy} = \frac{1}{N} \sum_{i=1}^{N} \text{I}(y_i = \hat{y}_i)
\end{equation}

where \( \text{I}(\cdot) \) is an indicator function that is $1$ if the condition is true and $0$ otherwise, \( y_i \) represents the true label, \( \hat{y}_i \) is the predicted label, and \( N \) is the total number of samples.

\begin{figure*}[!tb]
    \centering
    \begin{subfigure}[t]{0.32\textwidth}
        \includegraphics[width=\textwidth]{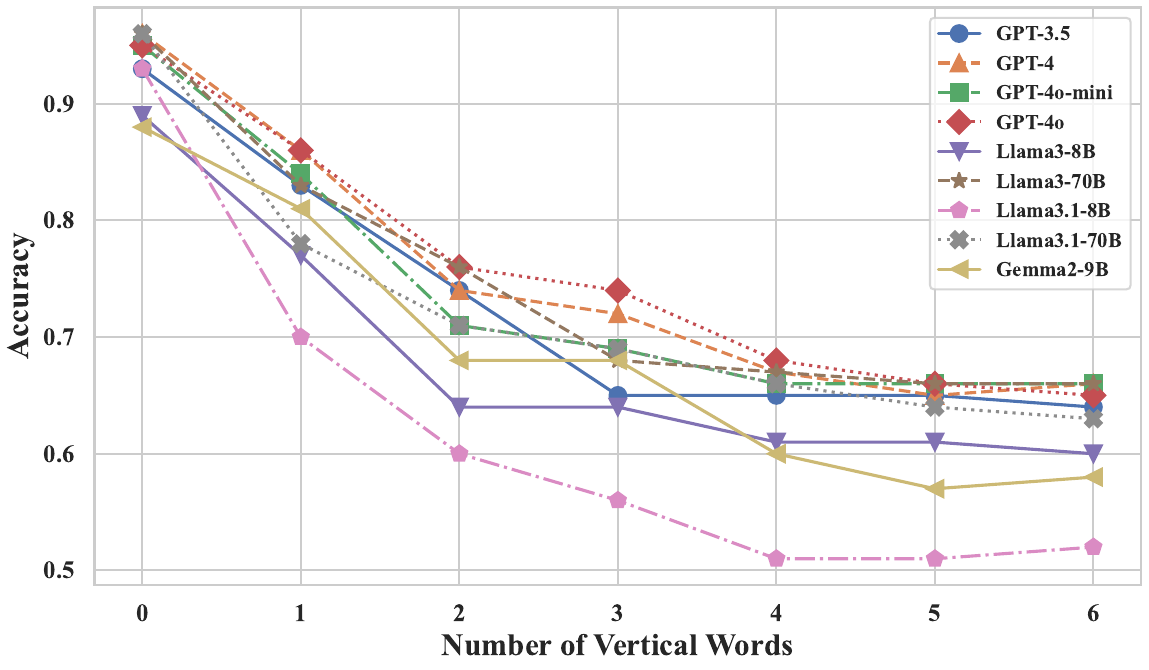}
        \caption{SST-2}
    \end{subfigure}    \hfill
    \begin{subfigure}[t]{0.32\textwidth}
        \includegraphics[width=\textwidth]{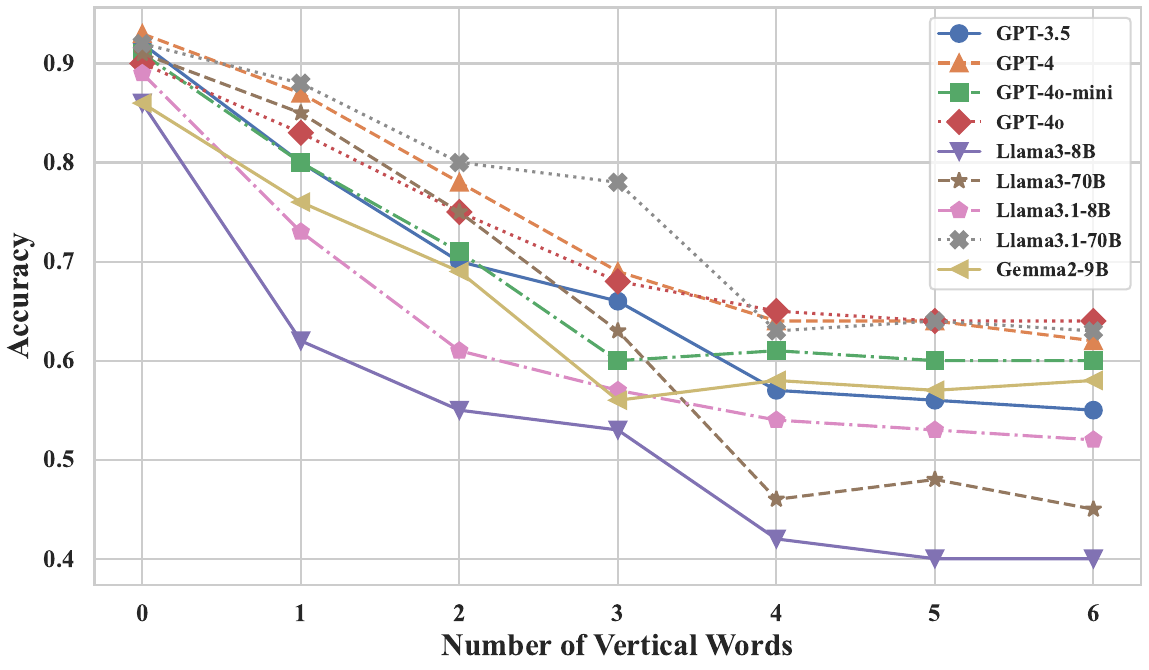}
        \caption{Rotten Tomatoes}
    \end{subfigure} \hfill
    \begin{subfigure}[t]{0.32\textwidth}
        \includegraphics[width=\textwidth]{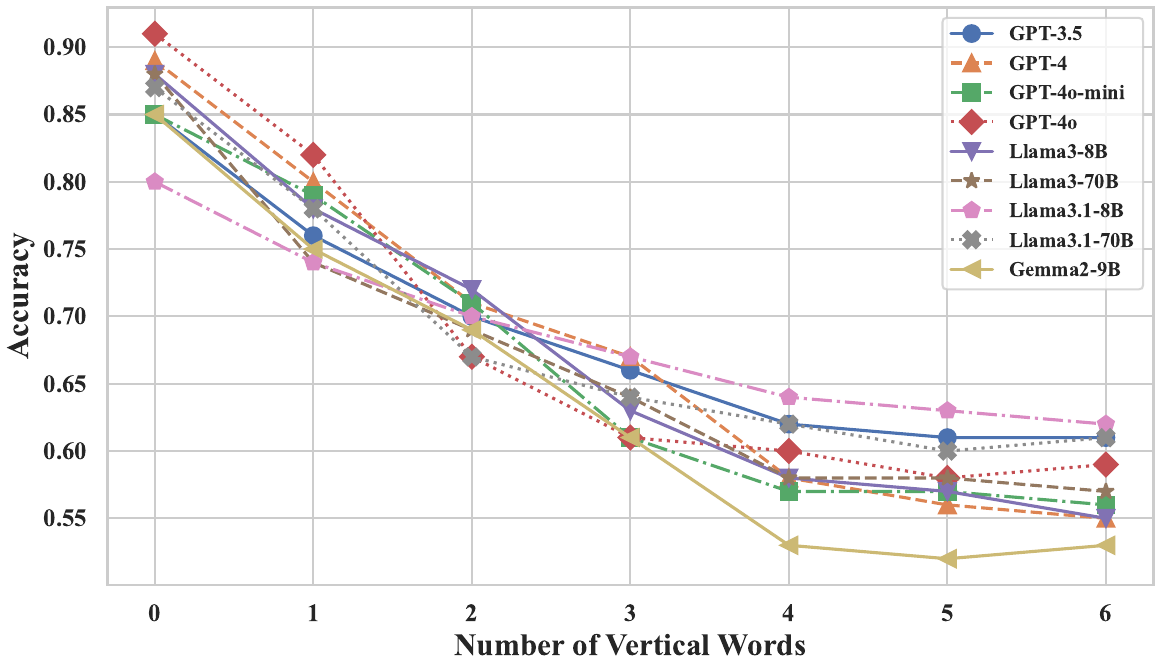}
        \caption{Jigsaw Toxicity}
    \end{subfigure}
    \vspace{-2mm}
    \caption{The relationship between the number of vertically inputted words and the accuracy of various LLMs on text classification tasks. \label{fig:wordnum}}
\end{figure*}

\subsection{Experiment Results}
We utilize four \gpt series LLMs along with eight widely recognized open-source LLMs on five text classification datasets. For the SST-2, QNLI, Rotten Tomatoes, and Jigsaw Toxicity datasets, the \gptfouromini model would select four words for vertical input. In contrast, for the CoLA dataset, it selects two words. 

In the label prediction phase, we utilize straightforward and precise prompts for zero-shot prediction, enabling the model to generate predicted labels. We compare the classification accuracy before and after modifying the input format, with the results presented in Table \ref{tab:mainexp}. The detailed implementation of different LLMs is shown in Appendix \ref{sec:implementdetail}.

\stitle{Vulnerability of LLMs to Vertical Input}
Based on the experimental results in Table \ref{tab:mainexp}, it is observed that inputting a few key words from the text in a vertical format into LLM significantly disrupts their ability to perform text classification tasks. 

Specifically, for the CoLA dataset, the model's classification accuracy drops by nearly $40$ percentage points when two key words are input vertically. Given that the random prediction accuracy for a binary classification problem is $50\%$, this indicates that the model essentially loses its classification ability. 

The model accuracy decreases the least on the QNLI dataset, likely due to the nature of the task. Relationships between sentences can be expressed in various ways, allowing LLMs to identify connections in the remaining content even when a few words are presented in a vertical format.

For the other three datasets, the decrease in accuracy after implementing vertical input for different models mostly ranges from $25$ to $40$ percentage points, which is a significant drop and almost indicates that current LLMs nearly lose their natural language understanding ability when encountering vertical input, even when the manipulation involves only a few key words.

\begin{table}[!tb]
    \footnotesize
    \begin{subtable}[h]{0.48\columnwidth}
    \centering
    \begin{tabular}{cc|c|c}
        \multicolumn{4}{c}{\textit{\textbf{Vertical}}}\\
         &  & \multicolumn{2}{c}{Actual} \\
         & & + & - \\\hline
         \parbox[t]{2mm}{\multirow{2}{*}{\rotatebox[origin=c]{90}{Pred.}}} & + & $14$ & $2$\\ 
         & - & $\textcolor{red}{36}$ & $48$ \\\hline
    \end{tabular}
    \end{subtable}
    \hfill
    \begin{subtable}{0.48\columnwidth}
        \centering
    \begin{tabular}{cc|c|c}
         \multicolumn{4}{c}{\textit{\textbf{Original}}}\\
         &  & \multicolumn{2}{c}{Actual} \\
         & & + & - \\\hline
         \parbox[t]{2mm}{\multirow{2}{*}{\rotatebox[origin=c]{90}{Pred.}}} & + & $43$ & $4$\\ 
         & - & $7$ & $46$ \\\hline
    \end{tabular}
    \end{subtable}
    \vspace{-2mm}
    \caption{Comparison of confusion matrices for the Jigsaw Toxicity dataset. (+) indicates toxic, while (-) indicates non-toxic.}
    \label{tab:jtcm}
\end{table}

\begin{table}[!tb]
    \footnotesize
    \begin{subtable}[h]{0.48\columnwidth}
    \centering
    \begin{tabular}{cc|c|c}
        \multicolumn{4}{c}{\textit{\textbf{Vertical}}}\\
         &  & \multicolumn{2}{c}{Actual} \\
         & & + & - \\\hline
         \parbox[t]{2mm}{\multirow{2}{*}{\rotatebox[origin=c]{90}{Pred.}}} & + & $39$ & $\textcolor{red}{41}$\\ 
         & - & $7$ & $13$ \\\hline
    \end{tabular}
    \end{subtable}
    \hfill
    \begin{subtable}{0.48\columnwidth}
        \centering
    \begin{tabular}{cc|c|c}
         \multicolumn{4}{c}{\textit{\textbf{Original}}}\\
         &  & \multicolumn{2}{c}{Actual} \\
         & & + & - \\\hline
         \parbox[t]{2mm}{\multirow{2}{*}{\rotatebox[origin=c]{90}{Pred.}}} & + & $43$ & $5$\\ 
         & - & $3$ & $49$ \\\hline
    \end{tabular}
    \end{subtable}
    \vspace{-2mm}
    \caption{Comparison of confusion matrices for the SST-2 dataset. (+) indicates positive and (-) indicates negative.}
    \label{tab:sst2cm}
\end{table}

\begin{table*}[!tb]
	\centering
	\begin{adjustbox}{width=1.00\linewidth}
    \renewcommand{\arraystretch}{1.20}
        \setlength{\tabcolsep}{19pt}
        \resizebox{\linewidth}{!}{
            \begin{tabular}{@{}lccccc@{}}
            \toprule
            \textbf{Model} & \textbf{SST-2} & \textbf{CoLA} & \textbf{QNLI} & \textbf{Rotten Tomatoes} & \textbf{Jigsaw Toxicity} \\ \midrule \midrule
            \multicolumn{6}{c}{\textbf{Closed-Source Models}} \\ \midrule
            \gptthreepointfive \,w/ CoT & $61.00$ $(\downarrow\textcolor{red}{4.00})$ & $50.00$ $(\uparrow\textcolor{darkgreen}{3.00})$ & $59.00$ $(\downarrow\textcolor{red}{10.00})$ & $53.00$ $(\downarrow\textcolor{red}{4.00})$ & $62.00$ $(0.00)$ \\
            \gptfour \,w/ CoT & $66.00$ $(\downarrow\textcolor{red}{1.00})$ & $51.00$ $(\uparrow\textcolor{darkgreen}{2.00})$ & $68.00$ $(\downarrow\textcolor{red}{3.00})$ & $60.00$ $(\downarrow\textcolor{red}{4.00})$ & $56.00$ $(\downarrow\textcolor{red}{2.00})$ \\
            \gptfouromini \,w/ CoT & $68.00$ $(\uparrow\textcolor{darkgreen}{2.00})$ & $50.00$ $(0.00)$ & $76.00$ $(\uparrow\textcolor{darkgreen}{1.00})$ & $64.00$ $(\uparrow\textcolor{darkgreen}{3.00})$ & $56.00$ $(\uparrow\textcolor{darkgreen}{3.00})$ \\
            \gptfouro \,w/ CoT & $71.00$ $(\uparrow\textcolor{darkgreen}{3.00})$ & $52.00$ $(\uparrow\textcolor{darkgreen}{5.00})$ & $74.00$ $(\uparrow\textcolor{darkgreen}{4.00})$ & $66.00$ $(\uparrow\textcolor{darkgreen}{1.00})$ & $66.00$ $(\uparrow\textcolor{darkgreen}{6.00})$ \\ \midrule
            \multicolumn{6}{c}{\textbf{Open-Source Models}} \\ \midrule
            \llamaone \,w/ CoT & $59.00$ $(\downarrow\textcolor{red}{2.00})$ & $49.00$ $(\downarrow\textcolor{red}{1.00})$ & $64.00$ $(\uparrow\textcolor{darkgreen}{2.00})$ & $53.00$ $(\uparrow\textcolor{darkgreen}{11.00})$ & $55.00$ $(\downarrow\textcolor{red}{3.00})$ \\ 
            \llamatwo \,w/ CoT & $66.00$ $(\downarrow\textcolor{red}{1.00})$ & $46.00$ $(\downarrow\textcolor{red}{2.00})$ & $59.00$ $(\downarrow\textcolor{red}{4.00})$ & $55.00$ $(\uparrow\textcolor{darkgreen}{9.00})$ & $62.00$ $(\uparrow\textcolor{darkgreen}{4.00})$ \\
            \llamathree \,w/ CoT & $53.00$ $(\uparrow\textcolor{darkgreen}{2.00})$ & $51.00$ $(\uparrow\textcolor{darkgreen}{2.00})$ & $62.00$ $(\uparrow\textcolor{darkgreen}{3.00})$ & $56.00$ $(\uparrow\textcolor{darkgreen}{2.00})$ & $63.00$ $(\downarrow\textcolor{red}{1.00})$ \\
            \llamafour \,w/ CoT & $68.00$ $(\uparrow\textcolor{darkgreen}{2.00})$ & $47.00$ $(\downarrow\textcolor{red}{3.00})$ & $65.00$ $(\downarrow\textcolor{red}{1.00})$ & $62.00$ $(\downarrow\textcolor{red}{1.00})$ & $65.00$ $(\uparrow\textcolor{darkgreen}{3.00})$ \\ \midrule
            \gemmaone \,w/ CoT & $60.00$ $(0.00)$ & $49.00$ $(\downarrow\textcolor{red}{2.00})$ & $64.00$ $(\uparrow\textcolor{darkgreen}{4.00})$ & $61.00$ $(\uparrow\textcolor{darkgreen}{3.00})$ & $52.00$ $(\downarrow\textcolor{red}{1.00})$ \\ 
            \gemmatwo \,w/ CoT & $61.00$ $(\uparrow\textcolor{darkgreen}{3.00})$ & $51.00$ $(\uparrow\textcolor{darkgreen}{1.00})$ & $64.00$ $(0.00)$ & $57.00$ $(\uparrow\textcolor{darkgreen}{3.00})$ & $53.00$ $(\downarrow\textcolor{red}{2.00})$ \\ \midrule
            \qwenone \,w/ CoT & $63.00$ $(0.00)$ & $51.00$ $(\downarrow\textcolor{red}{1.00})$ & $63.00$ $(\downarrow\textcolor{red}{3.00})$ & $59.00$ $(\uparrow\textcolor{darkgreen}{2.00})$ & $60.00$ $(\downarrow\textcolor{red}{3.00})$ \\
            \qwentwo \,w/ CoT & $62.00$ $(\uparrow\textcolor{darkgreen}{2.00})$ & $52.00$ $(\uparrow\textcolor{darkgreen}{2.00})$ & $64.00$ $(\uparrow\textcolor{darkgreen}{2.00})$ & $60.00$ $(\uparrow\textcolor{darkgreen}{1.00})$ & $61.00$ $(\uparrow\textcolor{darkgreen}{2.00})$ \\
            \bottomrule
            \end{tabular}
            \vspace{-2mm}
        }
	\end{adjustbox}
	\caption{Accuracy scores after applying CoT, with parentheses indicating the change in scores (compared to the non-CoT scores in Table \ref{tab:mainexp}). \textcolor{darkgreen}{Dark green} indicates an increase in accuracy, \textcolor{red}{red} denotes a decline, and \textcolor{black}{black} means no change. All other experimental settings remain consistent with those in Table \ref{tab:mainexp}.
    \label{tab:cotcompare}}
\end{table*}

\stitle{Potential Threat of Mislabelling}
According to Table \ref{tab:mainexp}, vertically formatted input text could reduce the accuracy of LLMs on sentiment and toxicity classification tasks by $30$ to $40$ percentage points. 
A more concerning issue emerges when examining the confusion matrix in Table \ref{tab:jtcm} and Table \ref{tab:sst2cm}: On the SST-2 dataset, inputting some words in a vertical format to the large language model reduces the classification accuracy for negative sentences from $91\%$ to $24\%$. Similarly, on the Jigsaw Toxicity dataset, this approach decreases the model's accuracy for classifying harmful text from $86\%$ to $28\%$.

This indicates that the models largely lose their ability to identify harmful content when key words are input vertically. In contrast, humans easily understand such text. Therefore, this vulnerability can have a severe negative impact on tasks like harmful content monitoring.

\stitle{Open-Source via Closed-Source}
In the upper part of Table \ref{tab:mainexp}, we evaluate four \gpt series models, considered some of the most powerful LLMs today. Regardless of model size and release date, they exhibit significant confusion when encountering vertically formatted text inputs. 

To ensure comprehensive experiments, we also test three high-performing open-source LLM series: \llama, \gemma, and \qwen. As shown in the lower part of Table \ref{tab:mainexp}, these open-source models also display considerable vulnerability to vertically formatted text inputs. Moreover, on four experimental datasets, \llamathree shows a greater accuracy drop than \llamaone, despite being a stronger language model. This suggests that current LLMs are not trained with consideration for the impact of vertical text inputs, resulting in a high degree of vulnerability, which may pose certain risks in some areas. 

% Moreover, on four datasets, \llamathree shows a greater accuracy drop than \llamaone, despite being a stronger language model. This suggests that understanding vertical input may not strongly correlate with benchmark scores.

\begin{figure*}[!tb]
	\centering
	\includegraphics[width=1.00\linewidth]{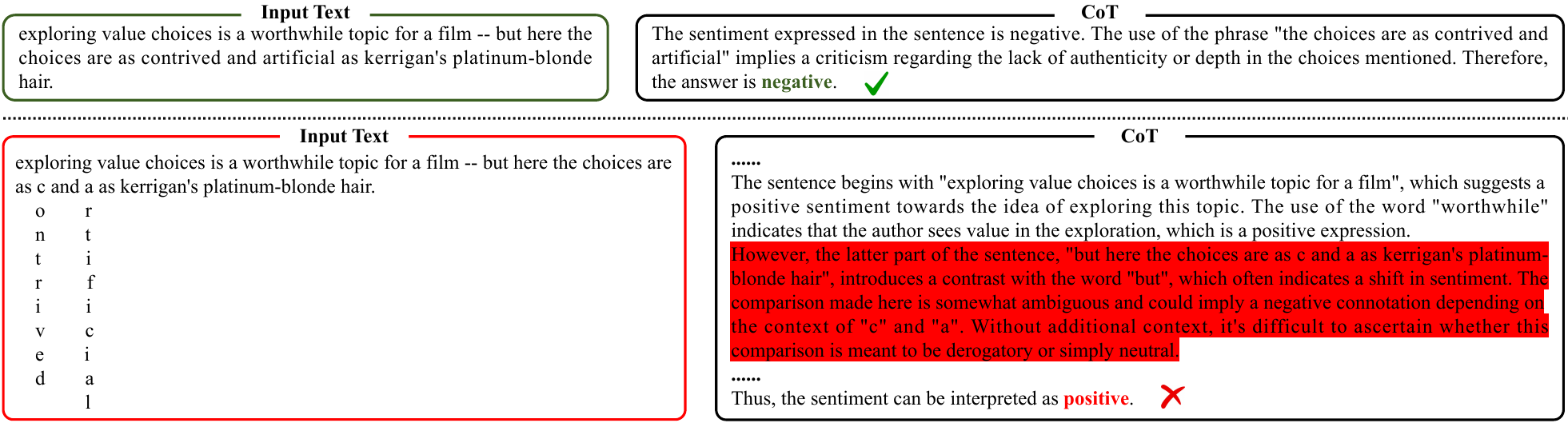}
	\caption{An example where Chain of Thought (CoT) reasoning is used during \gptfour inference, yet the text with some vertically input words is misclassified.
	\label{fig:cot}}
    \vspace{-2mm}
\end{figure*}

\stitle{Number of Vertical Words via Accuracy}
In Figure \ref{fig:wordnum}, we choose $9$ different LLMs to conduct a detailed experiment on the relationship between the number of vertically formatted words and the model's classification accuracy. 

Initially, as the number of vertically input words increases, the classification accuracy of the LLMs consistently declines. This happens because words relevant to the classification are gradually input in a vertical format, confusing the model's ability to distinguish them. When the number of vertically formatted words reaches a certain point, the model's accuracy stabilizes at a fixed value. This occurs because most of the key words affecting the classification are presented in a vertical format, and the remaining words have little impact on the model’s predictions. 

This experiment shows that the LLMs' classification heavily relies on the recognition of relevant words, further explaining its vulnerability to vertically formatted text. In contrast, identifying keywords related to sentiment or toxicity from a text is relatively simple for humans, highlighting the significant threat this vulnerability poses to LLMs.

\section{Potential Mitigation Methods}
\subsection{Failure of CoT}
In the above experiments, we have the LLMs directly output classification results for two types of text inputs. This method somewhat bypasses the model's reasoning process, which may result in inaccuracies. To address this, we consider using a Chain of Thought (CoT) approach~\citep{cot} to determine if prompting the model to output reasoning steps can mitigate its vulnerability to vertically formatted text inputs. We incorporate `think step by step' into the prompt to encourage the LLMs to provide a step-by-step reasoning process, aiming to help the model better understand the intricacies of vertically formatted text inputs~\citep{thinkstepbystep}.

However, the experimental results in Table \ref{tab:cotcompare} show that even with the use of the CoT, the LLMs still fail to recognize that some words are input in a vertical format, despite generating a corresponding reasoning process. 
We can observe that after using CoT with LLMs on different classification tasks, the classification accuracy either increases or decreases compared to not using it, which differs significantly from our expectations. Even when there is an improvement, the increase in accuracy is minimal, generally less than or equal to $3$ percentage points. On the Rotten Tomatoes dataset, although the two models from the \llamathreeseries series show a noticeable improvement of around $10$ percentage points after using CoT, their final classification accuracy remains extremely low, only about $55\%$. Unexpectedly, powerful models like \gptthreepointfive and \gptfour show reduced accuracy across four datasets despite using CoT.

\begin{figure*}[!tb]
    \centering
    \begin{subfigure}[t]{0.48\textwidth}
        \includegraphics[width=\textwidth]{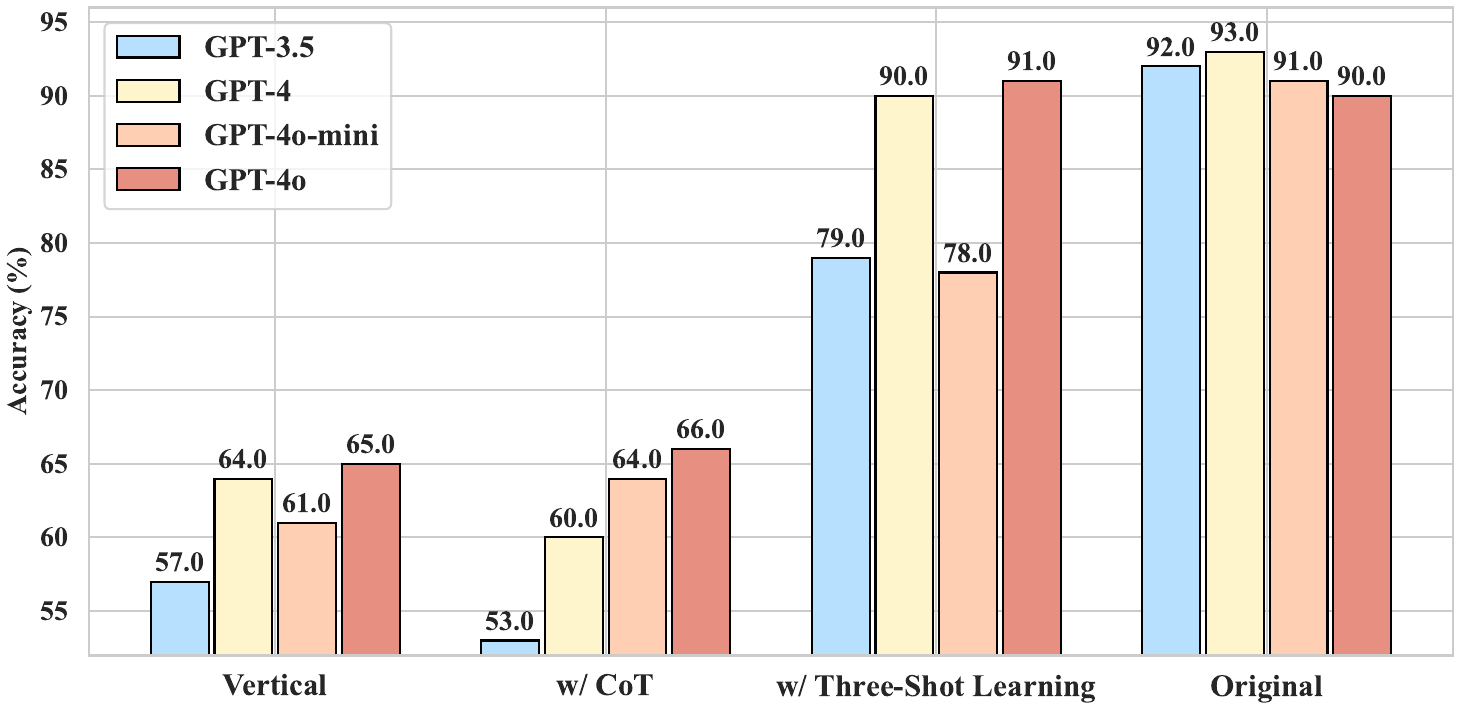}
        \caption{\textit{Rotten Tomatoes}}
    \end{subfigure} \hfill
    \begin{subfigure}[t]{0.48\textwidth}
        \includegraphics[width=\textwidth]{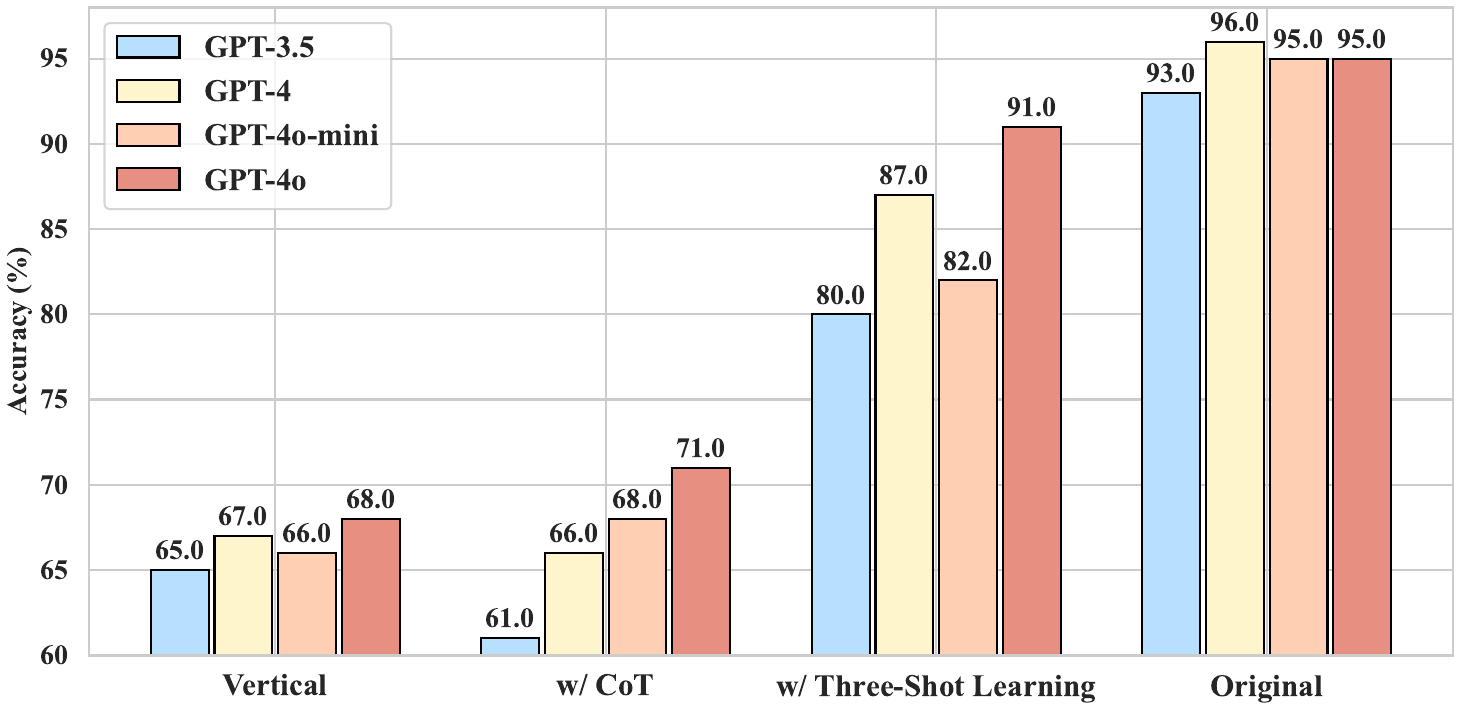}
        \caption{\textit{SST-2}}
    \end{subfigure}
    \caption{Comparison of accuracy scores between three-shot learning and three other methods.
    \label{fig:fewshot}}
    \vspace{-2mm}
\end{figure*}

Figure \ref{fig:cot} illustrates an instance where the powerful \gptfour model produces an incorrect answer despite utilizing the CoT method for reasoning. The model fails to associate the vertically formatted characters of each word, leading to guesswork and uncertainty, which results in the incorrect answer.

In summary, comparing the accuracy between the two types of inferences reveals no significant difference. This suggests that the chain of thought does not help the model reduce the interference caused by vertically formatted text inputs.

\begin{figure}
    \centering
\begin{tcolorbox}[top=2pt, bottom=5pt, colback=gray!10, boxrule=1pt, colframe=black, title=\gptfour Few-Shot Generation, fonttitle=\fontsize{10}{0}\selectfont, fontupper=\fontsize{9}{12.5}\selectfont]
To determine the sentiment of the sentence, let's first reconstruct it by identifying the vertically inputted words:

\greentext{\textbf{1.`overburdened' (o, v, e, r, b, u, r, d, e, n, e, d)}}

\greentext{\textbf{2.`complicated' (c, o, m, p, l, i, c, a, t, e, d)}}

\greentext{\textbf{3.`banal' (b, a, n, a, l)}}

Reconstructing the vertical part of the sentence: `overburdened with complicated plotting and banal dialogue'.

Now, let's analyze the sentiment of the reconstructed sentence:

(1) `overburdened' suggests something is excessively loaded or stressed.
(2) `complicated' implies something is overly complex.
(3) `banal' means something is unoriginal or boring.

Therefore, the overall sentiment is \greentext{\textbf{negative.} \, \checkmark}
\vspace{-2mm}
\end{tcolorbox}
    \caption{An example of few-shot learning output from the \gptfour model, in which the model identifies vertical words to arrive at the correct answer.}
    \label{fig:fewshotexample}
\end{figure}

\subsection{Effective Few-Shot Learning}
Our findings indicate that employing CoT does not effectively help LLMs identify words presented in vertical format, resulting in incorrect answers. 
We believe this limitation stems from the LLM's lack of awareness regarding such text formats. 

To address this, we propose using a few-shot learning approach to enhance the model's ability to recognize and respond to these special inputs. For each example, we meticulously construct a detailed analysis to facilitate the model's learning of relevant knowledge and assist in reconstructing the original sentence.

We employ a three-shot learning approach, selecting four models from the \gpt series for testing and validation across two datasets. As shown in Figure \ref{fig:fewshot}, the experimental results indicate a significant improvement in the model's accuracy after applying the three-shot learning method, especially with the \gptfour and \gptfouro models, which achieve prediction accuracy comparable to that of the original normal input text. This underscores the strong understanding and reasoning capabilities of these models. While the performance of \gptthreepointfive and \gptfouromini is lower than that of the first two models, they still demonstrate a noteworthy increase in accuracy. In Figure \ref{fig:fewshotexample}, the \gptfour model, using a three-shot learning method, accurately detects and reconstructs vertically formatted words from the input, generating correct classification outputs. This highlights the model’s enhanced performance with few-shot learning, particularly when compared to the CoT output in Figure \ref{fig:cot}.

\begin{figure*}[!tb]
    \centering
    \begin{subfigure}[t]{0.485\textwidth}
        \includegraphics[width=\textwidth]{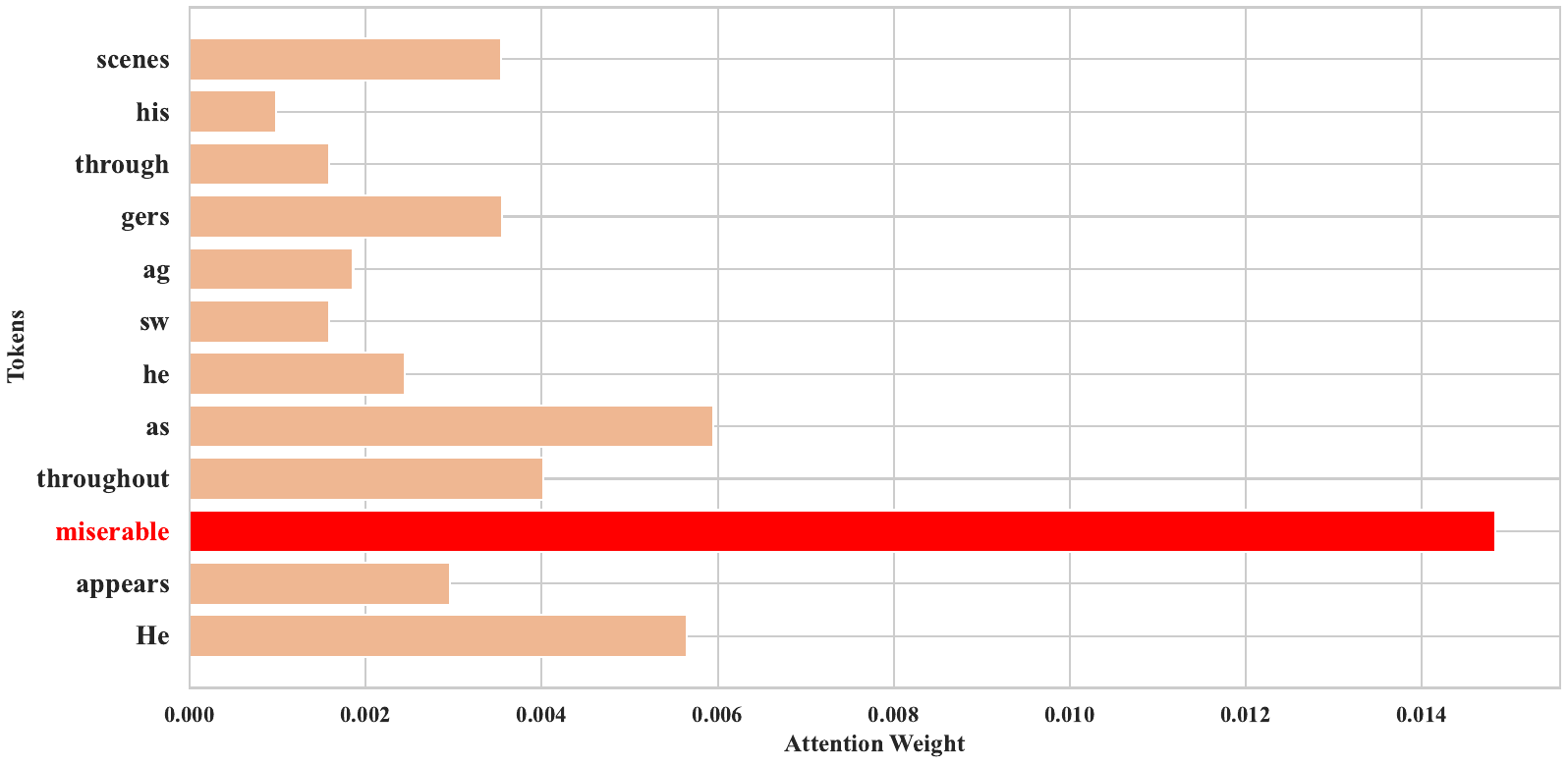}
        \caption{\textit{Original}}
    \end{subfigure} \hfill
    \begin{subfigure}[t]{0.485\textwidth}
        \includegraphics[width=\textwidth]{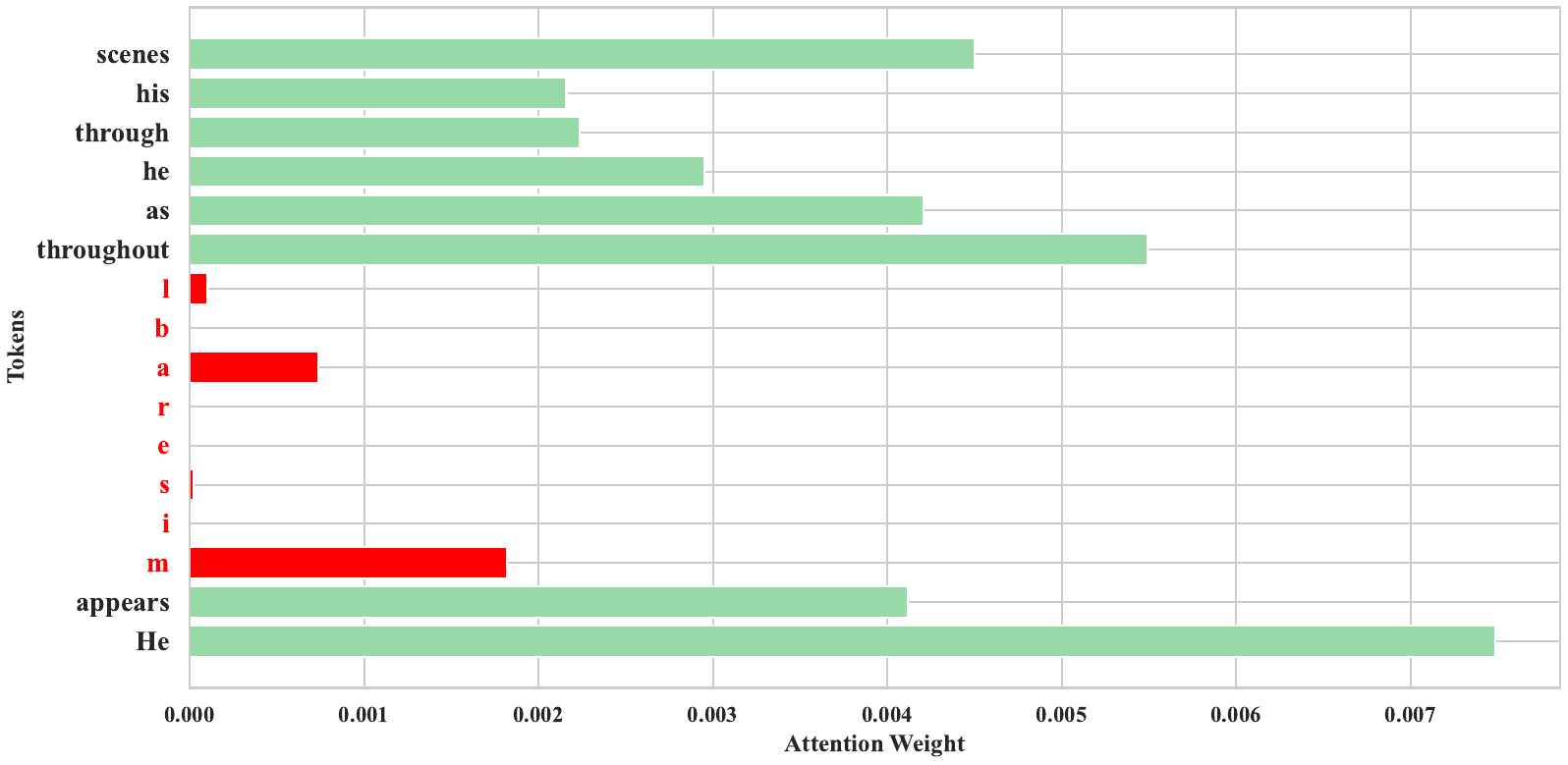}
        \caption{\textit{Vertical}}
    \end{subfigure}
    \caption{An example of comparing the attention weights of each token towards the `negative' token before and after vertical input of the word. The red highlights indicate the vertical word and its constituent tokens.
    \label{fig:attentioncompareone}}
    \vspace{-2mm}
\end{figure*}

\begin{figure*}[!tb]
    \centering
    \begin{subfigure}[t]{0.485\textwidth}
        \includegraphics[width=\textwidth]{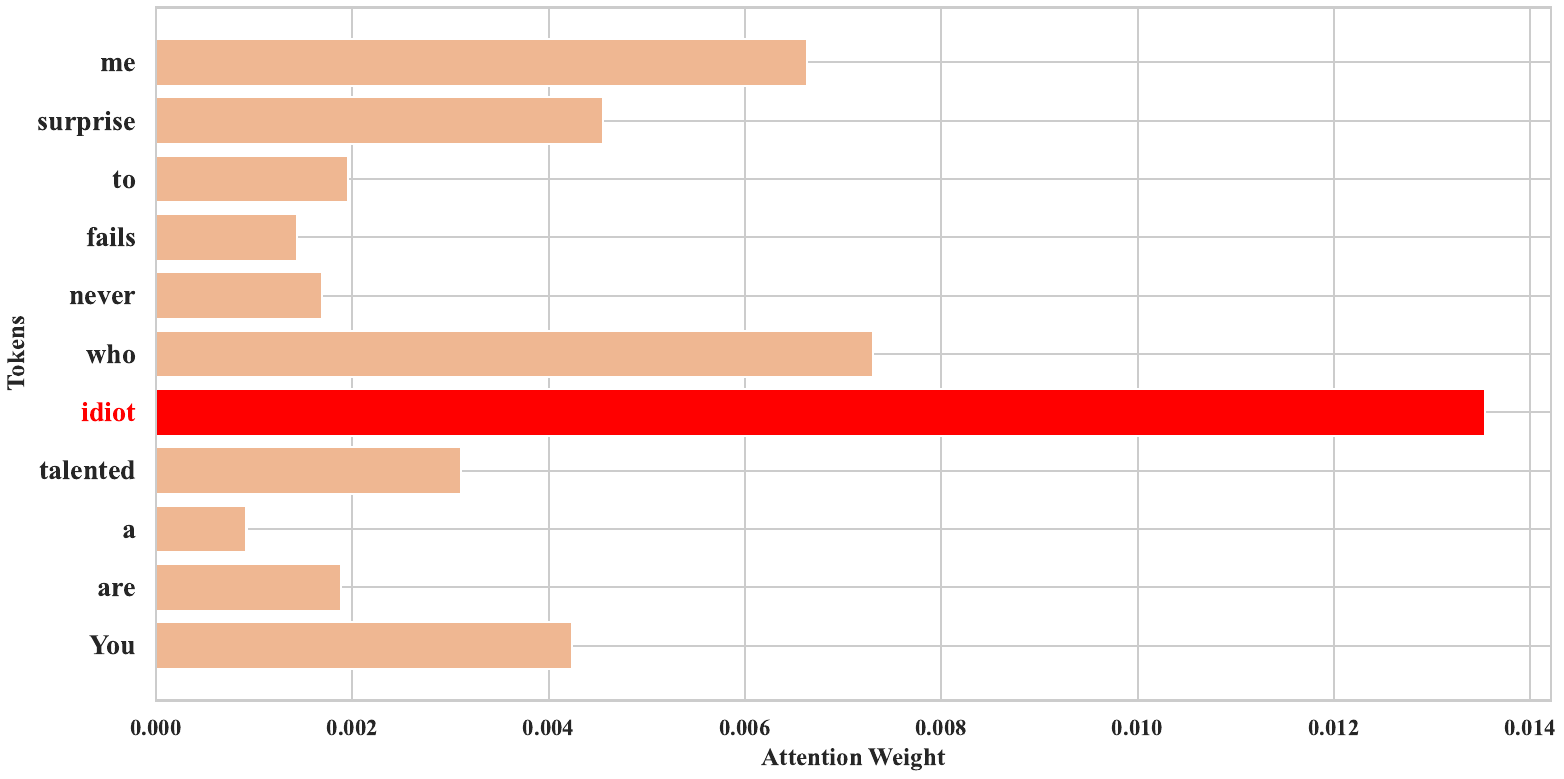}
        \caption{\textit{Original}}
    \end{subfigure} \hfill
    \begin{subfigure}[t]{0.485\textwidth}
        \includegraphics[width=\textwidth]{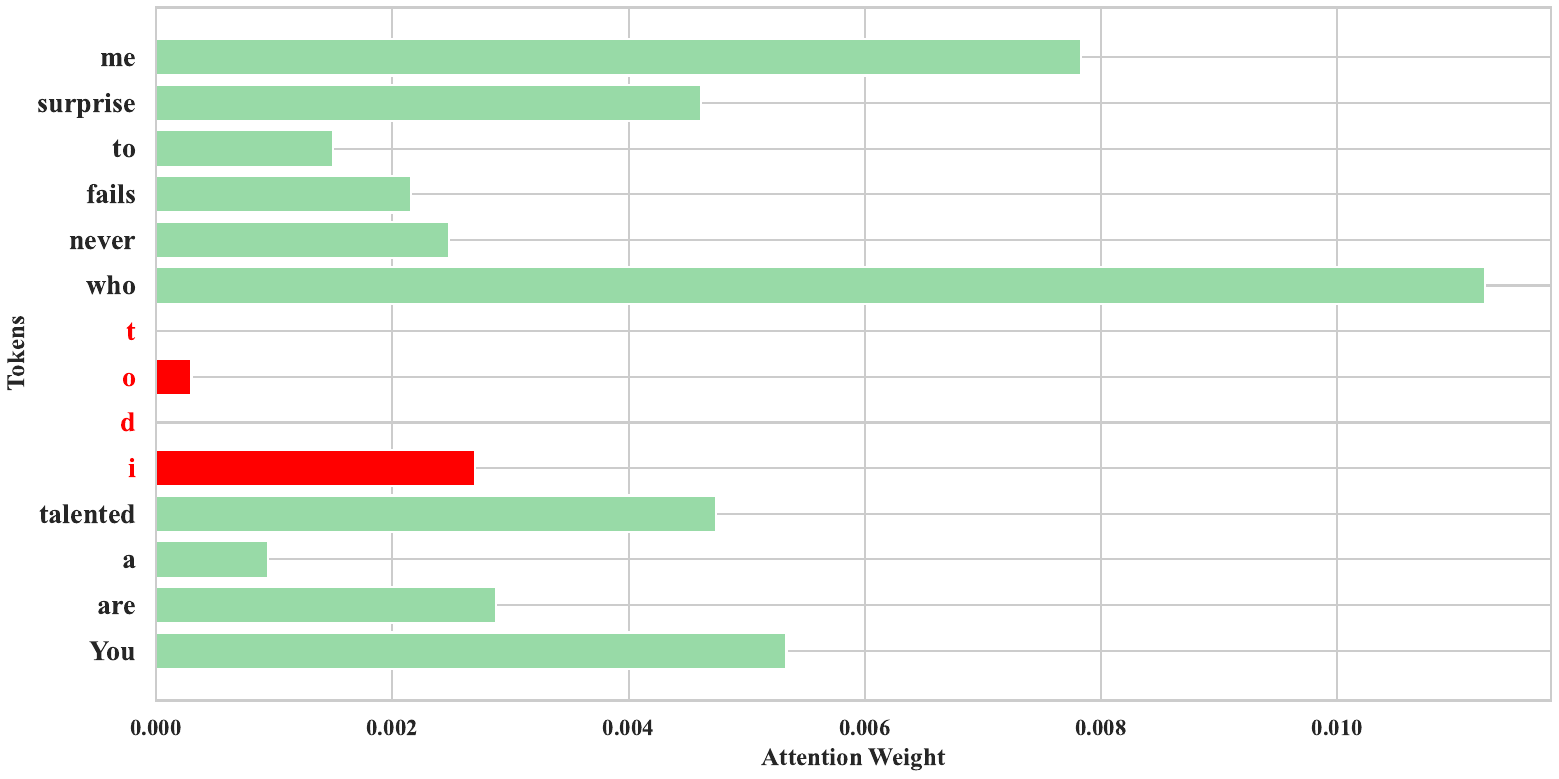}
        \caption{\textit{Vertical}}
    \end{subfigure}
    \caption{An example of comparing the attention weights of each token towards the `toxic' token before and after vertical input of the word. The red highlights indicate the vertical word and its constituent tokens.
    \label{fig:attentioncomparetwo}}
    \vspace{-2mm}
\end{figure*}

\section{Analysis}
\subsection{Disordered Tokens}
Due to the natural left-to-right and top-to-bottom writing order of text, the tokenizer used by LLMs encodes input text in this sequence. For instance, the tokenizer for the \llamathree model represents the word `vertical' as a single token when input horizontally. However, when input vertically, it converts into a token sequence of length 15 due to the spaces and line breaks. This causes the LLM to lose its understanding of the complete word.

\subsection{Lost Attention}
The most important part of LLMs is the attention matrix, which shows the degree of association between tokens within the model. We analyze the fundamental reason for the decrease in text classification ability of LLMs caused by vertical input text by plotting the changes in attention weights for corresponding words in the attention matrices of the \llamathree model, both before and after some words are input in a vertical format.

In Figure \ref{fig:attentioncompareone}, we use text from sentiment classification as an example, the original text is “He appears miserable throughout as he swaggers through his scenes”. We observe that when words are input into the large language model in a horizontal format, the model assigns strong attention weights to the word `miserable', which signifies a negative sentiment, in relation to `negative'. This indicates that the model recognizes `miserable' as conveying a negative emotion. However, when the word `miserable' is split and input in a vertical format, its components fail to establish a strong connection with `negative', ultimately leading to incorrect classification predictions by the model.

Similarly, in Figure \ref{fig:attentioncomparetwo}, the original toxic text is `You are a talented idiot who never fails to surprise me', when we input the harmful word `idiot' normally, the large language model connects `toxic' with `idiot' and classifies the text as harmful. However, when we input `idiot' in a vertical format, the model loses its ability to establish a strong connection between `toxic' and the new tokens (`i', `d', `o' and `t') that make up `idiot', leading it to conclude that the input is not harmful.

In summary, the tokenization conventions of LLMs 
% (processing text from left to right and top to bottom) 
and the lack of relevant pre-training data have impaired their ability to understand vertical text. In contrast, people can easily comprehend words presented in a vertical format. Consequently, the cognitive disparity between humans and LLMs leads to the models' vulnerability to vertical text input, which could pose potential threats that humans might exploit.

\section{Conclusion}
In this paper, we validate the vulnerability of modern LLMs to vertically formatted inputs, testing various mainstream models to reveal defects and flaws specific to this input type, which could pose real-world threats. 
Moreover, we find that CoT reasoning does not aid LLMs in resolving this issue, while few-shot learning with provided analysis could help mitigate it. We also note that this limitation arises from the nature of their pre-training data and tokenization mechanisms. In the future, we aim to explore more effective strategies for addressing this issue in LLMs through either pre-training or fine-tuning.

\section*{Limitations}
While we investigate the vulnerability of LLMs to vertically formatted input, this paper has several limitations:
(i) While our findings suggest that few-shot learning does help mitigate this vulnerability, it necessitates the design of demonstrations for each specific task. We do not explore the potential of fine-tuning LLMs to address this issue due to limitations in available datasets and GPU resources. Future research could investigate whether fine-tuning techniques could improve model robustness against vertically formatted input.
(ii) We do not assess the impact of vertically formatted input on text generation tasks. Future studies could investigate this aspect to evaluate any potential negative effects of such formatting on the generation performance of LLMs.

\section*{Ethics Statement}
Ethical considerations are of utmost importance in our research endeavors.  
In this paper, we strictly adhere to ethical principles by exclusively utilizing open-source datasets and employing various models that are either open-source or widely recognized in the scientific community.
Our findings highlight the text format vulnerabilities in large language models.
We are committed to upholding ethical standards throughout the research process, prioritizing transparency, and promoting the responsible use of technology for the betterment of society.
Additionally, we include a toxic example to highlight the potential severity of these vulnerabilities
To minimize negative impacts, we explore and provide a method for mitigation.

\section*{Acknowledgments}
This work is supported by NSF \#2331966, DARPA ANSR program FA8750-23-2-0004, an Amazon AGI Research Award, and University of California, Merced.
% The views and conclusions are those of the authors and should not reflect the official policy or position of DARPA or the U.S. Government.

\bibliography{custom}
\bibliographystyle{acl_natbib}

\newpage
\
\newpage
\appendix

\section{Large Language Models}
\label{sec:llmintro}
\begin{itemize}
\item {\gptthreepointfive}: A robust large language model developed by OpenAI, capable of generating text based on instructions, and highly effective across diverse natural language processing tasks.
\item {\gptfour}: An advanced multi-modal language model from OpenAI that accepts both image and text inputs for text generation, achieving near-human performance on various benchmarks.
\item {\gptfouromini}: A cost-efficient multimodal model, released by OpenAI on July 18, 2024, is a distilled version of \gptfouro, offering low latency and cost while supporting a wide range of tasks.
\item {\gptfouro}: OpenAI's latest multimodal AI model, offering enhanced reasoning, generation, and understanding capabilities across text, image, and speech with faster and more efficient responses.
\item {\llama}: Meta's open-weight AI models, designed for improved efficiency, reasoning, and multilingual capabilities, offering enhanced performance for various AI applications.
\item {\gemmatwoseries}: The next-generation open-source model from Google, released on June 27, 2024, as an improved version of {\gemma}, available in $2$B, $9$B, and $27$B parameter configurations.
\item {\qwen}: The series of models developed by Alibaba are powerful open-source AI models with strong multilingual understanding, containing different versions with varying parameters.
\end{itemize}

\section{Implementation Details}
\label{sec:implementdetail}
For the \gpt series models, we utilize the OpenAI API\footnote{\url{https://openai.com/}} for model invocation. During the word selection phase, we set \textit{top\_p} to $1.0$ and \textit{temperature} to $0.0$ to ensure consistent word selection. In the text classification phase, we adjust \textit{top\_p} to $0.95$ and keep \textit{temperature} at $0.0$ to maintain the reliability of the model's output.

For other open-source LLMs, we use either Hugging Face weights or the official API for model inference, applying the same parameter settings as above.

For each dataset, we randomly select $100$ test samples for experimentation. To ensure fairness, the number of samples for each label is evenly distributed in the selected test data, particularly for unbalanced datasets.

\section{Few-Shot with Open-Source LLMs}
\begin{figure}[!tb]
    \centering
    \includegraphics[width=1.00\linewidth]{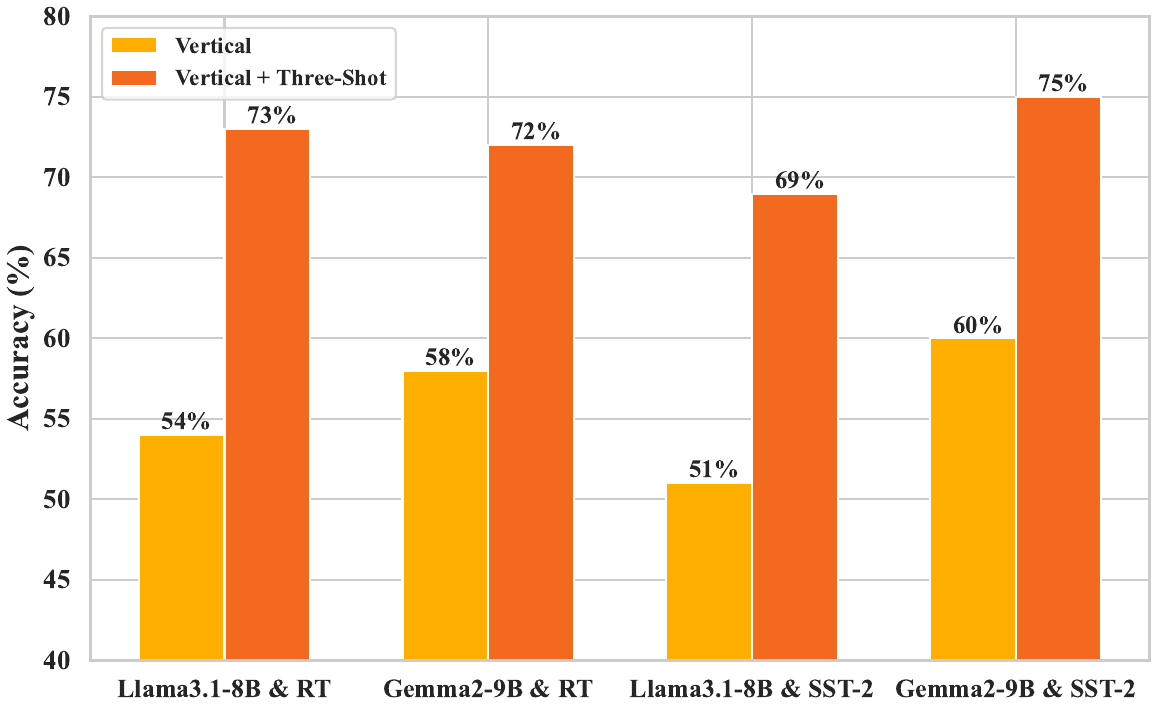}
    \caption{The accuracy comparison of two open-source LLMs after applying the three-shot learning method to interpret text input. The x-axis represents the model names and experimental datasets.
    \label{fig:appendixopensourcecomparison}}
    \vspace{-2mm}
\end{figure}

To further evaluate the effectiveness of the few-shot learning method in improving LLMs' ability to reconstruct vertical input content, we conduct experiments on two well-known open-source LLMs using two experimental datasets. The results are presented in Figure \ref{fig:appendixopensourcecomparison}.

Our findings show that applying three-shot learning significantly enhances model accuracy. \gemmaone achieves an improvement of approximately 19 percentage points across both datasets, while \llamathree gains around 15 percentage points. Although these improvements are smaller than those observed in the \gpt models shown in Figure \ref{fig:fewshot}, they still represent substantial gains, considering the base models' initial reasoning and natural language capabilities.

\section{Does Explicit Prompt Work?}
In real-world scenarios, natural system prompts do not explicitly instruct large language models to recognize and reconstruct vertical content in the input before generating the final output. To address this, we aim to examine how explicitly instructing the model to consider vertical content impacts its output.

In this section, we conduct an experiment by informing LLMs of the presence of vertical content in the input. Figure \ref{fig:explicitprompt} shows an example of an explicit instruction for the sentiment classification task.

\begin{figure}[!tb]
    \centering
    \includegraphics[width=1.00\linewidth]{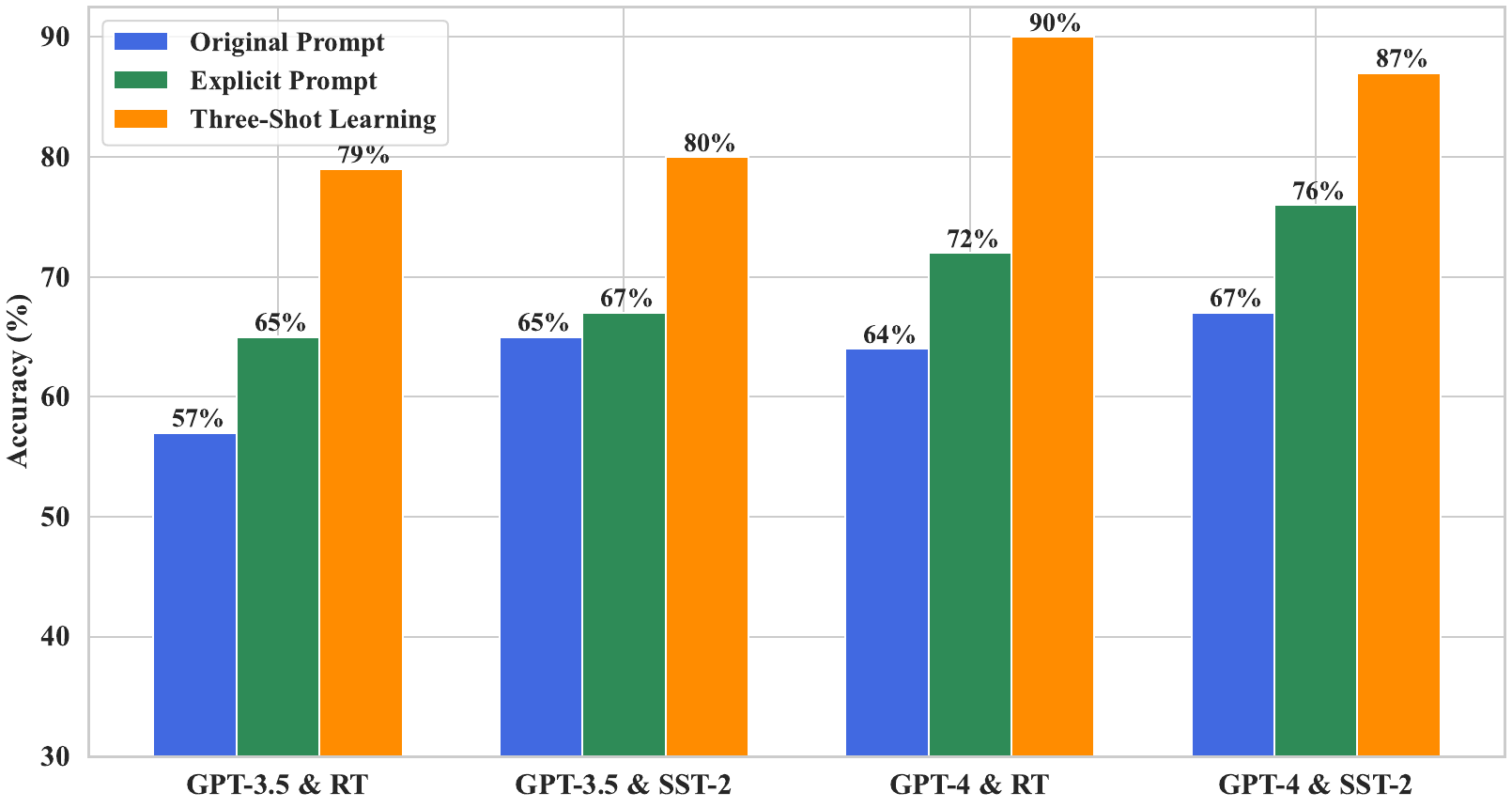}
    \caption{The accuracy comparison of two \gpt models under three different prompt settings, with the accuracy after using an explicit prompt shown in green. The x-axis represents the model names and experimental datasets.
    \label{fig:appendixpromptcomparison}}
    \vspace{-2mm}
\end{figure}

\begin{figure}
    \centering
\begin{tcolorbox}[top=5pt, bottom=12pt, colback=gray!10, boxrule=1pt, colframe=black, title=Explicit Prompt, fonttitle=\fontsize{10}{0}\selectfont, fontupper=\fontsize{10.5}{12.5}\selectfont]
I want you to determine whether the sentiment of a sentence is positive or negative. However, some of the words in the sentence are inputted vertically into the model. I expect you to identify these words, reconstruct the original sentence, and provide the correct sentiment classification.
\vspace{-2mm}
\end{tcolorbox}
    \caption{An example of an explicit instruction prompt for the sentiment classification task with vertical words input.}
    \label{fig:explicitprompt}
\end{figure}

% \textbf{\textit{I want you to determine whether the sentiment of a sentence is positive or negative. However, some of the words in the sentence are inputted vertically into the model. I expect you to identify these words, reconstruct the original sentence, and provide the correct sentiment classification.}}

We conduct the experiment using two \gpt models: \gptthreepointfive and \gptfour. We compare the performance of explicit prompts with naive and few-shot prompts to investigate the actual impact of explicit prompts in helping LLMs recognize vertical input. The results are shown in Figure \ref{fig:appendixpromptcomparison}.

The findings indicate that although explicit prompts help LLMs mitigate the negative impact caused by vertical input to some extent, the improvement is quite small compared to the few-shot prompt. On the SST-2 dataset, \gptthreepointfive improves its accuracy score by only $2$ percentage points with an explicit prompt, whereas it gains $15$ percentage points with a few-shot prompt. Similarly, for the \gptfour model on the Rotten Tomatoes dataset, the improvement is only $8$ percentage points with an explicit prompt, compared to $26$ with a few-shot prompt.

On the one hand, although explicit prompts improve model accuracy, the improvements are not substantial. On the other hand, this type of prompt is not widely applicable to real-world scenarios, as LLMs are exposed to numerous attacks and vulnerabilities.

\begin{table}[!tb]
	\centering
	\begin{adjustbox}{width=1.00\linewidth}
        \renewcommand{\arraystretch}{1.35}
        \setlength{\tabcolsep}{12pt}
        \resizebox{\linewidth}{!}{
    	\begin{tabular}{|c|c|c|}
            \toprule \toprule
            \textbf{Input Format} & \textbf{Dataset} & \textbf{Accuracy(\%)} \\ \midrule \midrule
            \multirow{5}{*}{\textit{Original}} & \textit{SST-2} & $96.67$ \\ \cmidrule{2-3} 
             & \textit{QNLI} & $92.00$ \\ \cmidrule{2-3} 
             & \textit{Rotten Tomatoes} & $97.33$ \\ \cmidrule{2-3} 
             & \textit{Jigsaw Toxicity} & $95.00$ \\ \midrule \midrule
            \multirow{5}{*}{\textit{Vertical}} & \textit{SST-2} & $96.33$ \\ \cmidrule{2-3} 
             & \textit{QNLI} & $92.00$ \\ \cmidrule{2-3} 
             & \textit{Rotten Tomatoes} & $97.33$ \\ \cmidrule{2-3} 
             & \textit{Jigsaw Toxicity} & $95.33$ \\ \bottomrule \bottomrule
            \end{tabular}
            \vspace{-2mm}
        }
	\end{adjustbox}
	\caption{The average accuracy of three human evaluations across four datasets, comparing results before and after vertical word input.
    \label{tab:human}}
\end{table}

\section{Human vs. LLMs}
To better understand the vulnerabilities of LLMs to vertical input compared to the human brain, we recruit three graduate-level students who are native English speakers to conduct experiments on four different datasets. Each student is presented with choices and required to select one, repeating this process twice: once before and once after the vertical input. Each time, the texts are shuffled to prevent memorization.

The experimental results of human classification are presented in Table \ref{tab:human}, with accuracy computed as the average performance of the three students for each task. The results indicate that vertical input poses no challenge to human cognition, as evidenced by a minimal accuracy difference of just $0.33\%$ across the two datasets, likely due to normal randomness rather than the input format. This demonstrates that while vertical input is easily comprehensible for humans, it remains a difficult problem for current LLMs, highlighting both the potential risks associated with relevant applications and the gap in format understanding between humans and LLMs.

\end{document}